\documentclass{article}

\PassOptionsToPackage{numbers,square}{natbib}

     \usepackage[preprint]{neurips_2024}

\usepackage[utf8]{inputenc} 
\usepackage[T1]{fontenc}    
\usepackage{hyperref}       
\usepackage{url}            
\usepackage{booktabs}       
\usepackage{amsfonts}       
\usepackage{nicefrac}       
\usepackage{microtype}      
\usepackage{xcolor}         
\usepackage{enumitem}
\usepackage{subcaption} 
\usepackage{bm}
\usepackage{graphicx}
\usepackage{amsmath}
\usepackage{listings}
\usepackage{xcolor}
\usepackage{multicol}

\definecolor{codegreen}{rgb}{0,0.6,0}
\definecolor{codegray}{rgb}{0.5,0.5,0.5}
\definecolor{codepurple}{rgb}{0.58,0,0.82}
\definecolor{backcolour}{rgb}{0.95,0.95,0.92}

\lstdefinestyle{mystyle}{
    backgroundcolor=\color{backcolour},   
    commentstyle=\color{codegreen},
    keywordstyle=\color{magenta},
    numberstyle=\tiny\color{codegray},
    stringstyle=\color{codepurple},
    basicstyle=\ttfamily\scriptsize,
    breakatwhitespace=false,         
    breaklines=true,                 
    captionpos=b,                    
    keepspaces=true,                 
    numbers=left,                    
    numbersep=5pt,                  
    showspaces=false,                
    showstringspaces=false,
    showtabs=false,                  
    tabsize=2
}
\lstdefinelanguage{yaml}{
  keywords={true,false,null,y,n},
  keywordstyle=\color{blue},
  basicstyle=\ttfamily\small,
  sensitive=false,
  comment=[l]{\#},
  commentstyle=\color{gray}\ttfamily,
  morestring=[b]',
  morestring=[b]",
  literate=
   *{:}{{{\color{black}:}}}{1}
    {-}{{{\color{black}-}}}{1}
    {>}{{{\color{black}>}}}{1}
    {|}{{{\color{black}|}}}{1}
}
\title{MXtalTools: A Toolkit for Machine Learning on Molecular Crystals}

\author{
Michael Kilgour$^{1}$\thanks{Corresponding author: \texttt{michael.kilgour@nyu.edu}}, \quad
Mark E. Tuckerman$^{1,2,3,4}$, \quad
Jutta Rogal$^{5}$\\[0.5em]
$^{1}$Department of Chemistry, New York University, New York, NY 10003, USA\\
$^{2}$Department of Physics, New York University, New York, NY 10003, USA\\
$^{3}$NYU–ECNU Center for Computational Chemistry at NYU Shanghai, Shanghai 200062, China\\
$^{4}$Simons Center for Computational Physical Chemistry at New York University, New York, NY 10003, USA\\
$^{5}$Initiative for Computational Catalysis, Flatiron Institute, New York, NY 10010, USA
}

\begin{document}

\maketitle

\begin{abstract}
We present MXtalTools, a flexible Python package for the data-driven modelling of molecular crystals, facilitating machine learning studies of the molecular solid state.
MXtalTools comprises several classes of utilities: (1) synthesis, collation, and curation of molecule and crystal datasets, (2) integrated workflows for model training and inference, (3) crystal parameterization and representation, (4) crystal structure sampling and optimization, (5) end-to-end differentiable crystal sampling, construction and analysis.
Our modular functions can be integrated into existing workflows or combined and used to build novel modelling pipelines.
MXtalTools leverages CUDA acceleration to enable high-throughput crystal modelling.
The Python code is available open-source on our \href{https://github.com/InfluenceFunctional/MXtalTools}{GitHub} page~\cite{kilgour2025mxtaltools}, with detailed documentation on \href{https://mxtaltools.readthedocs.io/en/master/index.html}{ReadTheDocs}~\cite{kilgour2025readthedocs}.
\end{abstract}

\section{\label{sec:Introduction}Introduction}

Molecular crystals are a diverse class of materials with many applications across numerous industries, including pharmaceuticals, agrochemicals, electronics, and energetic materials, and significant efforts have been devoted to understanding and engineering their properties.
While there is a large and mature software universe focused on data-driven modelling atomistic/inorganic materials, molecular crystals are distinct in important ways which call for purpose-built solutions.
For example, an accurate treatment of crystal symmetry operations applied to molecules is essential.

One of the primary computational challenges in the field of molecular crystals is crystal structure prediction (CSP), the determination of the most likely crystal structures for a given molecule, generally including structure generation, optimization, and ranking~\cite{beran2023frontiers}.
Related tasks, such as crystal representation, classification, and property prediction, are also of interest to many researchers.

Efficient machine learning on molecular crystals requires a set of minimal capabilities, including GPU acceleration, efficient batched operations, and support for automated gradient flow (autograd) through crystal building and analysis operations.
Existing packages for computationally-guided molecular crystal structure prediction~\cite{Tom2020,yang2025genarris, avery2017randspg, van1999upack, fredericks2021pyxtal}, while very useful, all lack one or more of these critical features, and hence are not suited to high throughput molecular crystals modelling.
MXtalTools (MXT) is an open-source Python code written specifically to address this shortfall, taking advantage of these capabilities to enable a wide range of modelling tasks, including crystal scoring, density prediction, optimization, and more.

A subtle yet important capability provided by native PyTorch~\cite{paszke2019pytorch} integration is that many of MXtalTools' workflows are end-to-end differentiable and fast enough to be included on-the-fly in training loops.
This enables immense flexibility in the types of workflows users can deploy, including crystal generation, optimization, or scoring, on any differentiably computable crystal or molecule property.

\section{\label{sec:Components}Components}

\begin{figure}
\centering
    \includegraphics[trim=0 160 0 160, clip, width=\textwidth]{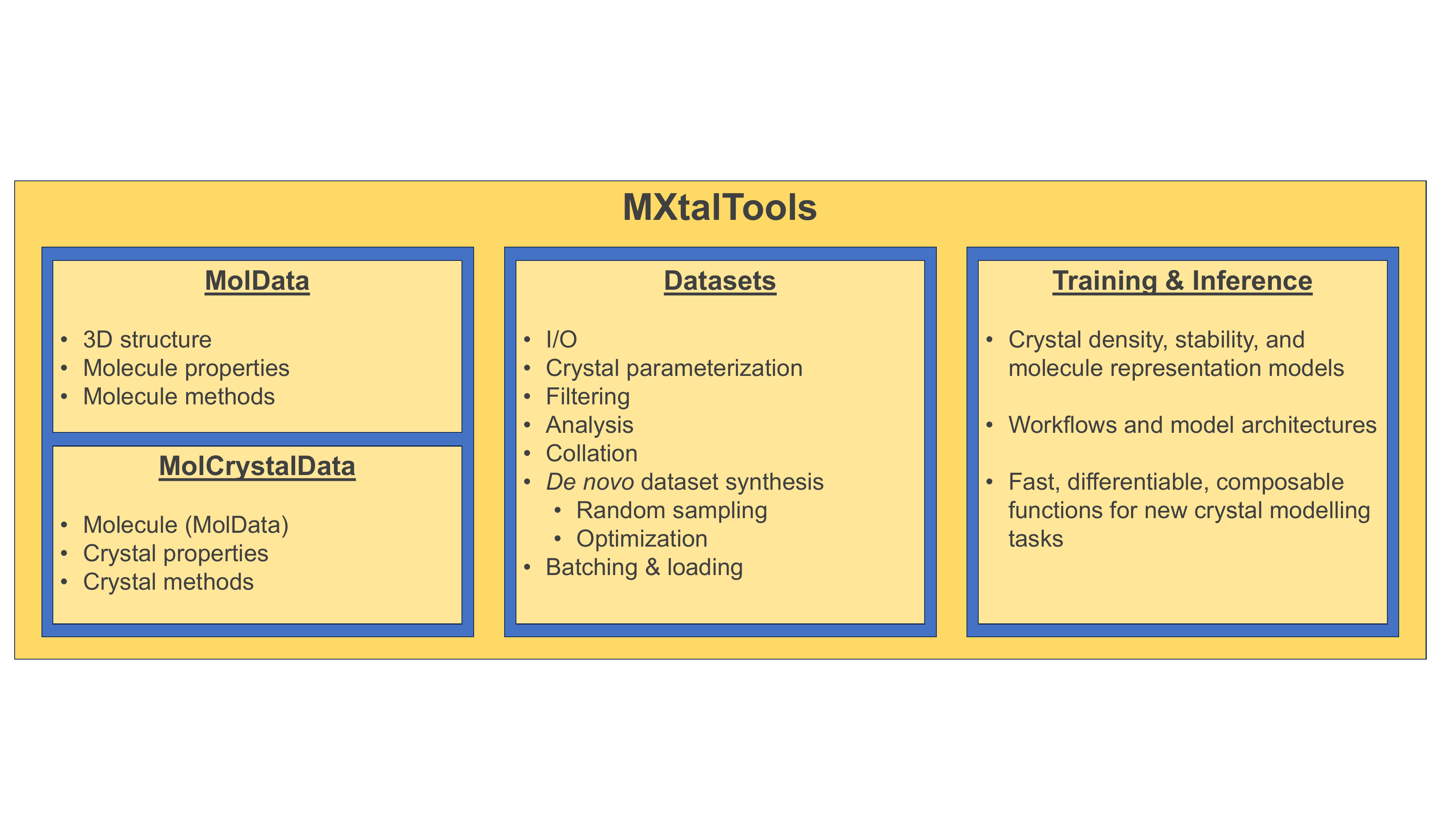}
    \caption{Major components of MXtaltools.}
\label{fig:components}
\end{figure}

The primary modules of MXT are summarized in Figure~\ref{fig:components}.
MXT is mainly structured around the construction and manipulation of molecule and crystal data objects, via \lstinline{MolData} and \lstinline{MolCrystalData} classes.
These data objects are created, parameterized, filtered, and collated through our database methods, for subsequent use in modelling workflows.

\subsection{\label{subsec:data_classes}Data Classes}
Inheriting batching and indexing behavior from the PyTorch Geometric \lstinline{Data} and \lstinline{Batch} classes, molecule and crystal objects can be collated and manipulated batch-wise for efficiency. 
Their instantiation and basic usage is illustrated with the following examples.

Molecules can be initialized either by explicit enumeration of the atoms, or built from SMILES strings~\cite{weininger1988smiles} (via RDKit~\cite{landrum2013rdkit}).
\begin{lstlisting}[language=Python] 
from mxtaltools.dataset_utils.data_classes import MolData

# initialize via atom types and coordinates
molecule = MolData(
    z=atomic_numbers_tensor,  # [n] 
    pos=coordinates_tensor,  # [n,3] 
    )

# initialize via SMILES string
base_molData = MolData()  # instantiate an empty class
molecule = base_molData.from_smiles( 
    smiles="CCC",  # a valid SMILES string
    minimize=True,  # optionally, minimize the 3D structure
    protonate=True,  # whether to include hydrogen
    )
\end{lstlisting}
\begin{figure}
\centering
    \includegraphics[trim=40 0 40 0, clip, width=\textwidth]{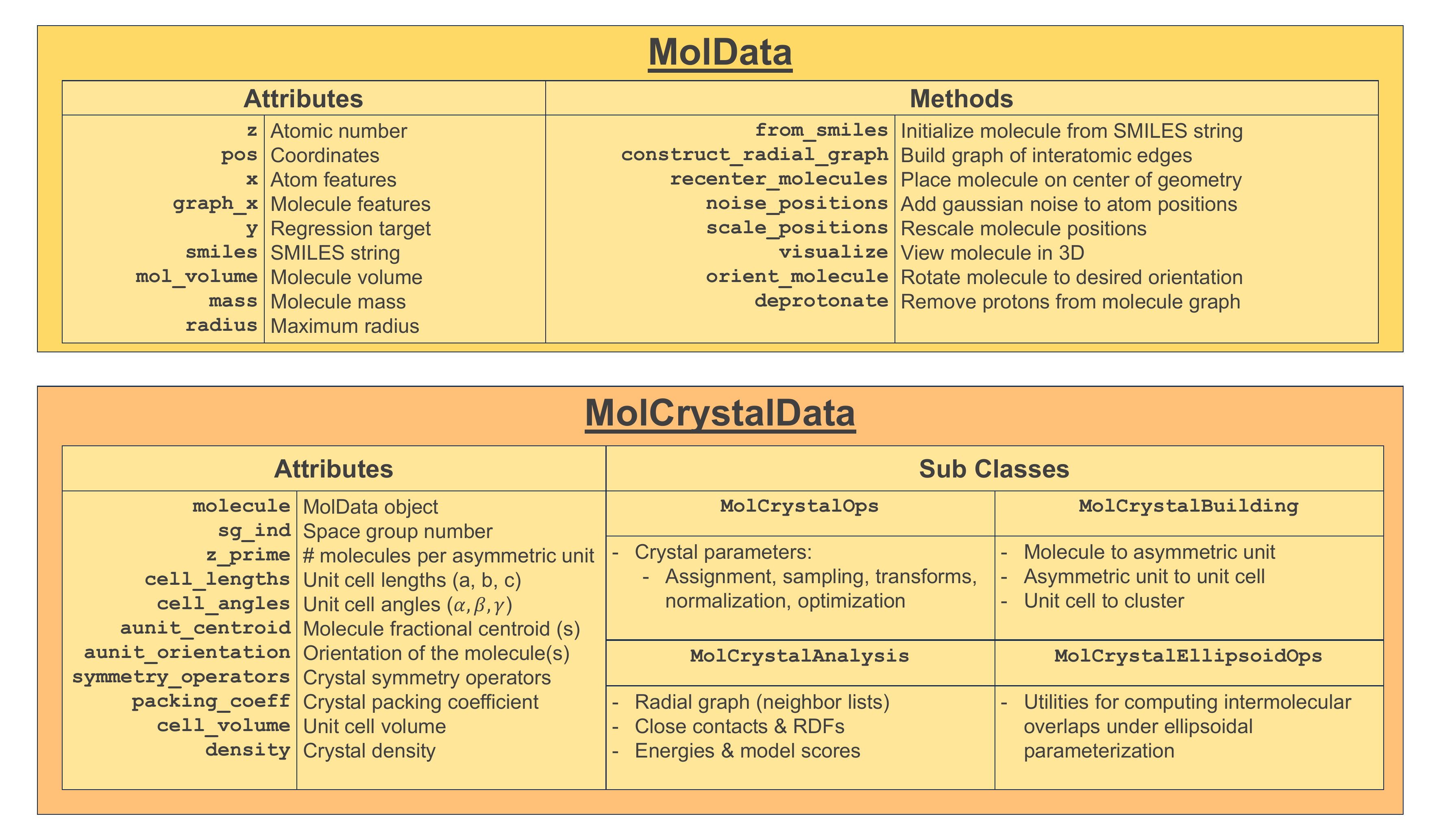}
    \caption{Class summary diagram for MolData and MolCrystalData objects.
    }
\label{fig:classes}
\end{figure}
Upon initialization, molecule geometric properties such as mass, volume, and radius are automatically computed.
Main attributes and methods are enumerated in Figure \ref{fig:classes}.

Molecular crystals are built from the combination of molecules, as instantiated above, and crystal parameters, defined in detail in the Supporting Information (SI). 
\lstinline{MolCrystalData} inherits class methods and properties from the original \lstinline{MolData}, and adds all the utilities required for parameterizing, constructing, and analyzing molecular crystals. 

\begin{lstlisting}[language=Python]
from mxtaltools.dataset_utils.data_classes import MolCrystalData

crystal = MolCrystalData(
    molecule,  # pre-instantiated MolData object
    space_group_number,  # integer (1-230) OR
    space_group_name,  # alternatively give the string representation
    z_prime,  # integer >= 1
    cell_lengths_tensor,  # [1,3] unit cell vector lengths, in Angstroms
    cell_angles_tensor,  # [1,3] interior cell angles, in radians
    aunit_centroid_tensor,  # [1,3*Z'] center-of-geometry of the molecule(s), in fractional coordinates of the unit cell
    aunit_orientation_tensor,  # [1,3*Z'] rotation vector to orient the molecule(s) in the asymmetric unit
    aunit_handedness_tensor,  # [1,1]  handedness of the molecule in the asymmetric unit
    )
\end{lstlisting}
Standard symmetry settings/operations are assumed, but can be overwritten when necessary.
Crystals are represented in a reduced-dimensional format via parameterization of molecules as rigid objects that are posed and oriented with respect to a `standard' initial state, defined as having the center of geometry at the origin and the molecule principal inertial axes aligned with the cartesian axes.
Combined with information on the box and symmetry information, these reduced degrees of freedom comprise the `crystal parameters'.
Important attributes and methods are shown in the class diagram in Figure \ref{fig:classes}.

Crystal parameters can be hard-coded, randomly generated, or extracted from unit cells of existing crystal samples (e.g., those extracted from databases).
We provide examples in Section \ref{sec:Examples}.

\subsection{\label{subsec:Dataset}Dataset Construction}

A typical MXtalTools training dataset is constructed by: 1) reading source files, typically crystallographic information files (cif's) for crystals and XYZ (Cartesian coordinate) files or SMILES for molecules, 2) computing the necessary sample features, (e.g., mass, volume, crystal parameters), and 3) instantiating each sample in a MolData or MolCrystalData object.
For crystal datasets, extensive filtering and processing is required to ensure data quality.
In common crystal datasets, such as the Cambridge Structural Database (CSD)~\cite{groom2016cambridge}, large numbers of crystal entries are either irrelevant for our purposes, such as disordered or polymeric structures, or have a serious flaw, such as missing atoms, or containing obviously unphysical structures.
The workflows and filtering conditions are detailed in the SI.

\subsection{\label{subsec:Training}Model Training}
We currently provide workflows to train models for molecule properties, crystal properties, molecule autoencoding, and molecule and crystal property prediction from fixed embeddings (see workflow diagrams in the SI).
We have trained models for crystal density (equivalent to unit cell volume, crystal packing coefficient), given only the molecule graph, as well as the twenty QM9/QM9s~\cite{ramakrishnan2014quantum,zou2023deep} quantum mechanical properties of small organic molecules.
Additional property prediction models are easily trainable, provided sufficiently large synthetic or experimental training datasets.
These could include material properties such as the Young's modulus, brittleness, plasticity and elasticity, for which datasets could plausibly be generated with accurate general potentials~\cite{wood2025family,gharakhanyan2025open} and appropriate simulation protocols~\cite{elgengehi2025accurate}.
We plan also to train a model for high precision molecular volume prediction in the near future.

Given the modular nature of our training and reporting utilities, adding new workflows is straightforward.
Training runs are controlled via .yaml configuration files, and logged to the Weights \& Biases online platform through our \lstinline{Logger} class, which also collects and processes convergence and evaluation statistics. 

\subsection{\label{subsec:Models}Models}

Included under the \verb|models| directory is a library of custom machine learning models, modules, and utilities.
Through a combination of standard scalar/invariant neural network operations and O(3) equivariant operations, as detailed in~\cite{kilgour2024multi}, we construct:
\begin{enumerate}
\item Standard multilayer perceptrons (MLP, or feedforward fully connected neural networks), with skip connections, normalization, and dropout options.
\item Equivariant MLP models (EMLP), for modelling higher-order tensor properties such as dipole or quadrupole moments.
\item Standard and equivariant graph neural networks (GNNs), correspondingly modelling scalars and vectors, respectively.
The GNN models can be used on single molecules or clusters carved out of molecular crystal supercells. 
The clusters represent the crystalline environments required in the computation of the embedding for the canonical conformer in the molecular crystal graph~\cite{kilgour2023}. 
\end{enumerate}
A typical crystal scalar property prediction model might comprise 4 graph convolution layers and 4 MLP layers, all with 512 feature channels, layer normalization, and dropout probability of 0.25.

All our graph models are built on a \lstinline{BaseGraphModel} class, which assumes an input in the form of a point cloud with atom types and positions, and optionally molecule-wide properties, and can be mixed and matched with MLP models as appropriate for a given modelling task.

\section{\label{sec:Examples}Examples}

In this section, we will demonstrate the deployment of some of our utilities and pretrained models on practical analyses.
As of time of writing, the crystal volume and score models are trained on the $Z'=1$ CSD structures, excluding polymers, porous structures, nonstandard symmetries, and erroneous structures, as outlined in \cite{kilgour2023}.
The Mo3ENet molecule autoencoder~\cite{kilgour2024multi} was trained on randomly generated conformers of `QM9-like' molecules from the QM9 and ZINC22 datasets~\cite{ruddigkeit2012enumeration,ramakrishnan2014quantum,tingle2023zinc}, meaning they contain H, C, N, O and F, with up to 9 heavy atoms.

\subsection{\label{subsec:example_autoencoder} Molecule Representation}

As a step towards modelling molecular crystals via a low-dimensional embedding rather than with an all-atom approach, we developed the Mo3ENet (molecular O(3) equivariant encoder net) autoencoder model~\cite{kilgour2024multi}.
Mo3ENet converts molecule point clouds into representations that are equivariant to rotation and inversion, and that provably contain complete information on atom positions and types. 

Generating a molecule embedding is done by passing a batch of molecules to the autoencoder model, and retrieving the equivariant `vector' embedding, and its rotationally invariant sub-representation, the `scalar' embedding.
\begin{lstlisting}[language=Python]
from mxtaltools.dataset_utils.utils import collate_data_list
from mxtaltools.common.training_utils import load_molecule_autoencoder

device = 'cuda'  # select device 'cuda' or 'cpu'
MolData_list = torch.load(molecules_path, weights_only=False)  # load list of pre-instantiated molecules
mol_batch = collate_data_list(MolData_list)  # collate into batch object
mol_batch.to(device)  # pass to relevant device

model = load_molecule_autoencoder(checkpoint, device)  # load model

molecule_batch.recenter_molecules()  # center molecules on (0,0,0)
vector_encoding = model.encode(molecule_batch.clone())  # get vector embedding
scalar_encoding = model.scalarizer(vector_encoding)  # get scalar embedding

# confirm reconstruction quality
reconstruction_loss, mean_deviation, matched_molecule = \
    model.check_embedding_quality(molecule_batch, visualize=False)
    
\end{lstlisting}
We can confirm the embedding is of sufficiently high quality by computing its reconstruction loss and mean deviation from the target structure, and even visualize the molecule and reconstruction together.
If the reconstruction is accurate, the embedding necessarily contains complete information about the molecular point cloud.

\subsection{\label{subsec:example_density} Crystal Density Prediction}
Following our work in~\cite{kilgour2023}, we illustrate a workflow for predicting the unit cell volume for a given molecule, without any direct knowledge of the crystal structure.
The quantity to be regressed can be cast in several equivalent ways: crystal packing coefficient, $C_P=\frac{V_\text{mol}\cdot Z}{V_\text{cell}}$ where $Z$ is the symmetry multiplicity of the space group (number of molecules per unit cell), crystal density $\rho$, or asymmetric or unit cell volume, $V_\text{aunit}, V_\text{cell}$.
The crystal packing coefficient is generally preferred as it is unitless and largely independent of molecule size and space group.
There is no universally accepted definition of molecule volume, and as such no general method to compute it.
To get consistent volume estimates from given $C_p$ model, one must therefore use the molecule volume calculator used when preparing the training data.

Running inference with a pretrained model is straightforward and the corresponding workflow would be very similar for any molecule property prediction task.
One has only to load and batch the molecules, load the model, and then evaluate the desired property:
\begin{lstlisting}[language=Python]
from mxtaltools.common.training_utils import load_molecule_scalar_regressor
from mxtaltools.dataset_utils.utils import collate_data_list

device = 'cuda'  # select device 'cuda' or 'cpu'
MolData_list = torch.load(molecules_path, weights_only=False)  # load list of pre-instantiated molecules
mol_batch = collate_data_list(MolData_list)  # collate into batch object
mol_batch.to(device)  # pass to relevant device

model = load_molecule_scalar_regressor(model_checkpoint_path, device)  # load model

output = model(mol_batch).flatten()  # call model

# final answer and optional unit conversions
packing_coefficient = output * model.target_std + model.target_mean  # destandardize
asymmetric_unit_volume = mol_batch.mol_volume / packing_coefficient  #  A^3
density = mol_batch.mass / asymmetric_unit_volume * 1.6654  # g/cm^3
\end{lstlisting}

\subsection{\label{subsec:example_crystal} Crystal Building and Analysis}

A core component of our codebase is the construction and analysis of molecular crystals. 
Most straightforwardly, one may load prebuilt crystals, build explicit convolution clusters, and analyze them:

\begin{lstlisting}[language=Python]
from mxtaltools.models.utils import softmax_and_score
from mxtaltools.dataset_utils.utils import collate_data_list
from torch.nn.functional import softplus
from mxtaltools.common.training_utils import load_crystal_score_model

device = 'cuda'  # select device 'cuda' or 'cpu'
MolCrystalData_list = torch.load(crystals_path, weights_only=False)  # load list of pre-instantiated crystals
crystal_batch = collate_data_list(MolCrystalData_list)  # collate into batch object

model = load_crystal_score_model(checkpoint, device)  # load model

cutoff = 6  # six Angstrom intermolecular edge cutoff

# construct explicit unit cell and convolution cluster
cluster_batch = crystal_batch.mol2cluster( 
    cutoff=cutoff,  
    align_to_standardized_orientation=True,  # build from standardized initial pose
    )

cluster_batch.construct_radial_graph(cutoff=cutoff)  # get intermolecular edges

model_output = model(cluster_batch)  # consult crystal score model
# model output has shape [n_samples, 3]
# first two output dimensions are probability amplitudes for (fake, real)
classification_score = softmax_and_score(model_output[:, :2])
# third dimension gives the predicted earth mover's distance to the ground truth in RDF space
predicted_RDF_distance = softplus(model_output[:, 2])

# we can also compute simple intermolecular potentials, such as Lennard Jones or short-range electrostatics
outputs = cluster_batch.compute(computes=['lj','es','silu'])

# visualize crystal structures with ASE
indices_to_visualize = [1, 2, 3, 4, 5]  # look at crystals 1-5
cluster_batch.visualize(
    indices_to_visualize, 
    mode='convolve with'  # visualize molecules that are inside the convolution window
    )
\end{lstlisting}

The \lstinline{MolCrystalData} interface also allows for easy crystal analysis, including Lennard-Jones energies, a short-range electrostatic potential, and a repulsion-softened Lennard-Jones-style energy (`SiLU' potential, see definition in SI).
We also provide interfaces for the popular MACE and UMA machine learned interatomic potentials (MLIPs)~\cite{Batatia2022mace,wood2025family,gharakhanyan2025open}, and can easily include additional MLIPs with pre-existing Atomic Simulation Environment (ASE)~\cite{larsen2017atomic} interfaces.
Sample visualization is handled via the use the ASE package's \lstinline{visualize.view} function.

Our crystal score models (MXtalNet-S) are trained on experimental crystal structures from the CSD and randomly generated crystal `fakes'.
MXtalNet-S predicts two values: 1) a classification score, representing the confidence the crystal \textbf{is} the experimentally observed structure, and 2) a distance metric quantifying the distance in radial distribution space between a given crystal and the experimental structure.
This radial distribution function earth mover's distance (RDF EMD) is defined in detail in the Supplemental Material (SI).

Our crystal building and analysis functions are all differentiable, enabling the computation of gradients through crystal analysis and construction, back to the crystal parameters or even a generating function for said parameters, by calling \lstinline{output.mean().backward()} on the outputs,.
We demonstrate a practical application in Section \ref{sec:case_study}.

\section{\label{sec:case_study} Case Study}

\begin{figure}
\centering
    \includegraphics[trim=0 220 0 200, clip, width=\textwidth]{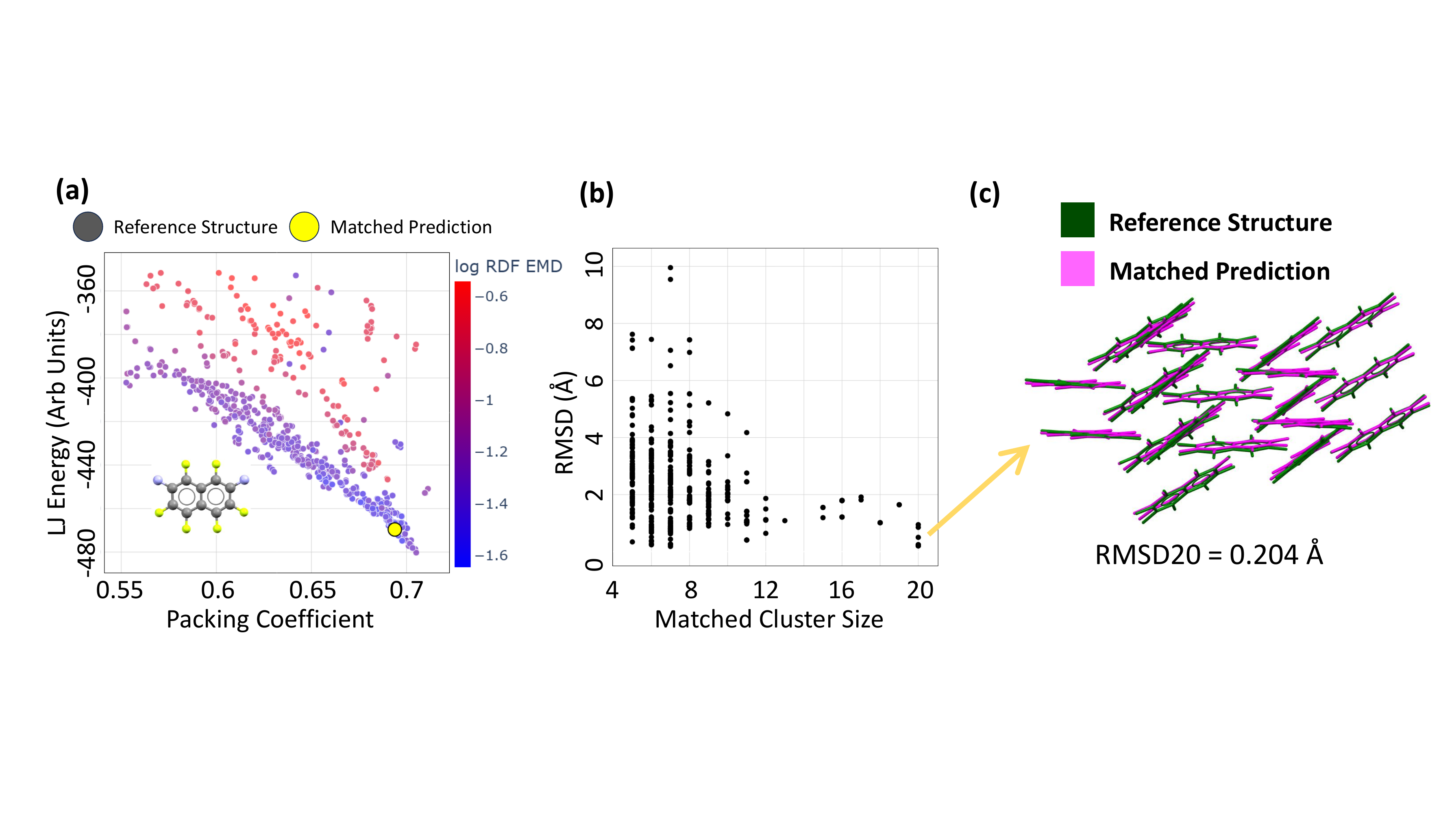}
    \caption{(a) Results of a crystal search run, showing the intermolecular LJ energy as a function of density (crystal packing coefficient), colors indicate the log of the RDF EMD between each sample and the experimental crystal structure, 
    (b) RMSD and number of matched molecules for 755 structures,
    (c) the closest 20/20 matched cluster with RMSD of 0.204\AA. 
    }
\label{fig:compack}
\end{figure}

To show how one can combine MXtalTools modules into useful workflows, we present a case study of a basic crystal structure prediction.
The goal of this study is to identify likely crystal structures of a given molecule, 1,3,4,5,6,8-hexafluoronaphthalene-2,7-diamine, CSD ID: DAFMUV (inset in Figure \ref{fig:compack} panel (a)).
DAFMUV has an experimentally known crystal structure~\cite{vaganova2017design} in space group number 33: Pna21, serving as a reference for the predicted structures.
Because our workflows operate on rigid molecules and DAFMUV is itself fairly rigid, we use the experimental conformer directly as input.

We leverage the end-to-end differentiability afforded by PyTorch integration, and optimize the crystal parameters directly via backpropagation through molecule posing, crystal building, and scoring/analysis.
Our crystal search module can optimize crystals on any differentiable PyTorch function, from simple energies to pretrained crystal score models such as MXtalNet-S.
For this case study, we opted for simple energy functions to showcase the entire workflow with minimal computational cost.

We begin our search (procedure detailed in the SI) by estimating the likely density of the crystal using MXTalNet-D, following the example workflow in~\ref{subsec:example_density}, and receive an estimated crystal packing coefficient of \textbf{0.699}, within 1\% agreement with the reference value of \textbf{0.694}.

Samples are initialized for optimization with random crystal parameters, and densities set at the target value.
Random sampling of molecular crystal parameters for realistic densities typically yields crystals with intermolecular overlap.
The too short distances between molecules result in extremely high energy `jammed' structures, which often fail to optimize to reasonable local minima.

We therefore optimize in two stages, first based on a soft-repulsion potential (details in the SI), the squared deviation of the density from the predicted value, and an auxiliary loss to prevent cell vectors from getting too long.
The softened interatomic repulsion allows molecules with severe overlap to slide past one another, while still equilibrating to a minimally overlapping local minimum. 
In the second stage, a short local refinement with respect to the LJ energy and the target density is performed.
For simplicity, we omit further refinements, but users can replace the LJ energy or do subsequent optimizations on desired energy functions, including MXtalNet-D, or pretrained MLIPs.

Panel (a) in Figure~\ref{fig:compack} shows the energy vs. density plot for a batch of 940 optimized crystal structures, with 60 very poor quality structures excluded from an original 1000.
The search found a number of structures near the expected experimental density with comparable and even lower energies than the experimental structure. 
To determine if these lower energy structures are truly more stable than the experimental one, structural refinement and energy evaluations with more accurate energy functions would be needed.
Structures similar in densities and energy to the experimental target structure have usually, though not always, the smallest EMD in RDF space, indicating their packing patterns are 
in good agreement with the experimental one.

In panel (b) of Figure~\ref{fig:compack}, we show the root mean square deviation (RMSD) of the generated structures to the experimental one is shown for 755 out of 1000 structures, focusing on samples with packing coefficients between 0.6-0.8. 
RMSD values and number of matched molecules were computed using the COMPACK algorithm~\cite{chisholm2005compack} (details in the SI).
The COMPACK analysis confirms that our search yielded many structures with good local structural agreement with the experimental polymorph.
In panel (c), we visualize the closest match, with an RMSD20 of 0.204\AA. 
The Lennard-Jones potential seems to prefer slightly higher-densities than the experimental structure, yet within 1000 generated samples we were able to find five 20/20 cluster matches. 

This case study and the examples in Section~\ref{sec:Examples} demonstrate complicated multi-step workflows that can be easily performed with just a few MXtalTools functions.
Improvements or modularization can be incorporated seamlessly, for example, a generative model could be substituted for random sampling for initial candidate structures, genetic optimization, Monte Carlo, or other optimization schemes could be combined with gradient descent, and any learned model or autograd-friendly property could be used for scoring and ranking.
This flexibility and amenability to both novel experimentation and efficient implementation are the core strengths of this codebase.

\section{\label{sec:Summary}Summary and Outlook}

\subsection{Use Cases}
We have introduced our software package, MXtalTools, containing several utilities for the construction, analysis, and modelling of molecular crystals. 
Particularly time-saving are our crystal dataset curation pipeline, our crystal building and analysis methods, and our automated training and inference workflows. 
The modules form the core utilities for MXT users when developing their own tools and methods.
MXT's general end-to-end differentiability also offers promising opportunities for novel crystal sampling and optimization approaches.

Our tools provide significant support for researchers pursuing a variety of objectives, for example:
\begin{enumerate}
    \item Experimentalists or computational chemists can easily install the MXtalTools package and estimate the density of a given molecular crystal, based on their molecule(s) of study.
    \item Molecular ML scientists can integrate our Mo3ENet molecular embedding models into their workflows, using the equivariant representation as a stand-in for molecules themselves.
    \item Computational molecular crystals researchers can use our analysis tools and crystal scoring models to analyze and optimize their candidate structures.
    \item Molecular crystal ML researchers can mix and match our modular toolkit to construct custom workflows for novel modelling tasks, including backpropagation through all steps of crystal construction and modelling.
\end{enumerate}

\subsection{Limitations and Future Directions}
MXtalTools currently supports homomolecular (identical molecules and conformers) crystals, for integer $Z'$, and general Wyckoff positions.
Molecules are generally assumed to be rigid, though torsional flexibility may be added in a future update.
To explore different conformers, one must reinitialize MolData objects with those conformations directly.
Unique asymmetric unit parameterization is currently only possible in space groups where the asymmetric unit is an easily defined parallelepiped.
This includes most space groups below 99 and a heterogeneous set above.
The relevant groups are explicitly defined in \verb|constants/asymmetric_units.py|.

Extensions to all space groups, explicitly flexible molecules, and cocrystals, are mostly technically straightforward and planned for the future implementations.
Integration of further new models, including generative models for crystals, will come as such new capabilities are developed. 

\textbf{Data and Software Availability.} With the exception of Cambridge Crystallographic Data Centre (CCDC)~\cite{Sykes:oc5038} software, all the software components required to run MXtalTools are available free of charge, under BSD-3 license. 
Detailed installation instructions are available on our \href{https://github.com/InfluenceFunctional/MXtalTools}{GitHub} README page~\cite{kilgour2025mxtaltools}.
Some crystal processing functions currently require the CCDC Python API, that requires an active paid license to run.
The Cambridge Structural Database ~\cite{groom2016cambridge} likewise requires a paid license to access in bulk.
MXtalTools is tested on Linux and Windows with Python $\geq$ 3.10.

\section*{Acknowledgements}
JR acknowledges financial support from the Deutsche Forschungsgemeinschaft (DFG) through the Heisenberg Programme project 428315600. MK, JR, and MET acknowledge funding from grants from the National Science Foundation, DMR-2118890, and MET from CHE-1955381. This work was supported in part through the NYU IT High Performance Computing resources, services, and staff expertise.
The Flatiron Institute is a division of the Simons foundation.

\bibliography{main}

\renewcommand{\thefigure}{S\arabic{figure}}
\renewcommand{\thetable}{S\arabic{table}}
\renewcommand{\theequation}{S\arabic{equation}}

\renewcommand{\thesection}{S\arabic{section}}  
\setcounter{section}{0}  

\section{RDF Distance Calculation}
For the estimation of crystal similarity, we compute for each sample the radial distribution (RDF) from 0-6 \r{A}, for all unique pairs of atom types (elements) in the crystal.
These pairwise RDFs describe the local neighborhood about a given molecule.

To garner a usable metric from these local `fingerprints', we compute the earth mover's distance between pairs of crystals' RDFs, and sum the element pair contributions according to their probability mass.
This corresponds qualitatively to the distance atoms would need to move in order to transform from one crystal to another. 

Stepwise, we compute the RDF distance between crystal $1$ and $2$ for element pair $(i,j)$ (for example, carbon and nitrogen), via the 1D earth movers distance along the radial direction, with $N$ discrete bins $r$,
\begin{equation}
    EMD_{pair}(\mathcal{R}_{1}^{ij}, \mathcal{R}_{2}^{ij}) = \frac{6}{N^2}\sum_{k=0}^N\left|\sum_{r=0}^k\mathcal{R}_1^{ij}(r)-\sum_{r=0}^k\mathcal{R}_2^{ij}(r)\right|,
\end{equation}
with $\mathcal{R}^{ij}_m$ the discrete radial distribution function for crystal $m$, and atom pair $i$ and $j$.
We take this distance for all pairs of elements in each crystal, and average the them according to their relative total RDF mass,
\begin{equation}
    EMD_{tot}(\mathcal{R}_1, \mathcal{R}_2) = \frac{1}{n}\left(\sum_{ij}^nw_{ij}EMD_{pair}(\mathcal{R}_{1}^{ij}, \mathcal{R}_{2}^{ij})\right),
\end{equation}
for
\begin{equation}
    w_{ij} = \frac{1}{2}\left(\sum_{r=0}^N\mathcal{R}_1^{ij}(r)+\sum_{r=0}^N\mathcal{R}_2^{ij}(r)\right),
\end{equation}
for all $n$ unique atom pairs, $ij$.

\section{Workflows}
\begin{figure}
\centering
    \includegraphics[trim=40 0 40 0, clip, width=\textwidth]{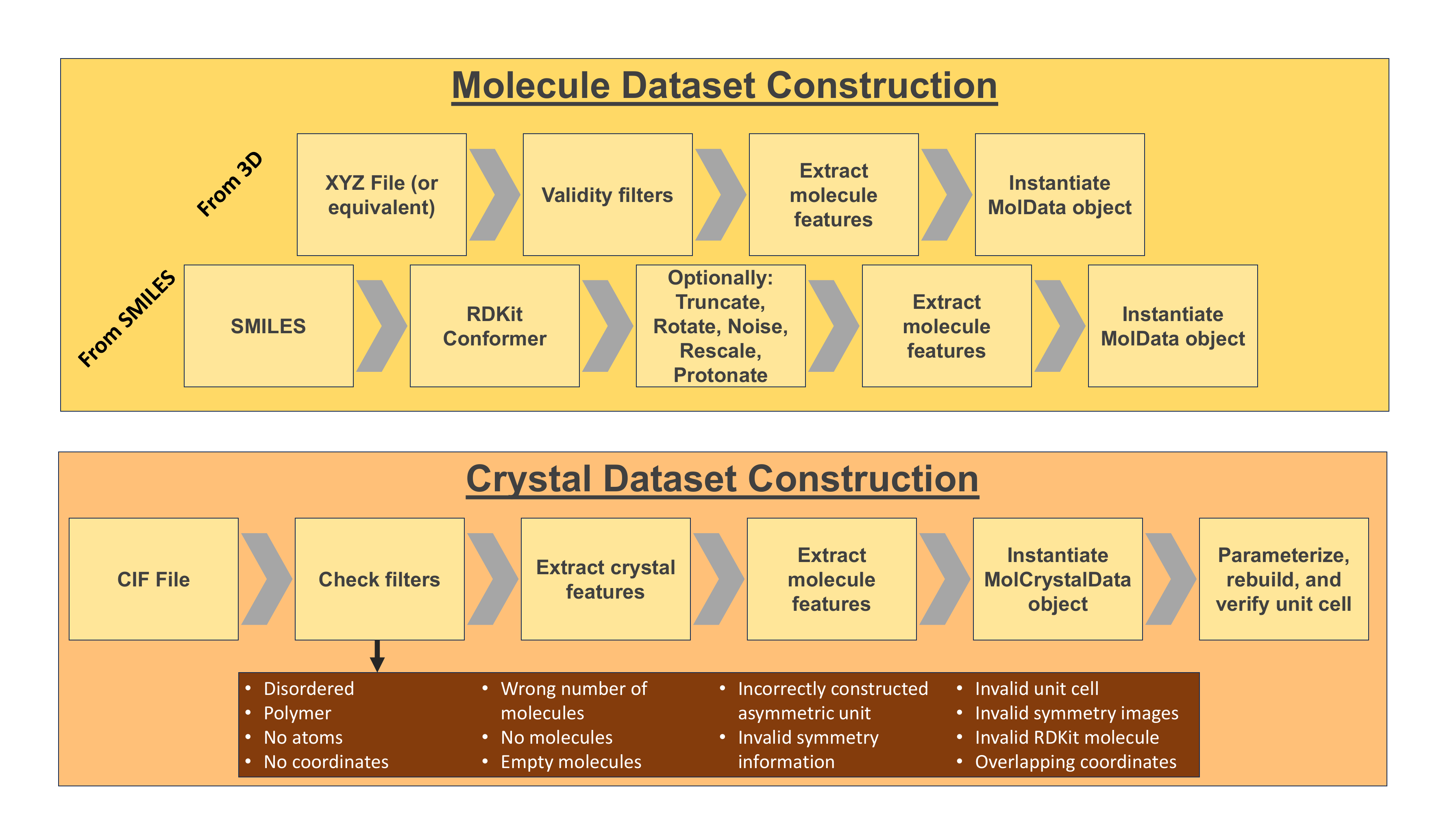}
    \caption{Workflows for molecule and crystal data point creation.
    }
\label{fig:mol_workflow}
\end{figure}

\begin{figure}
\centering
    \includegraphics[trim=60 180 60 180, clip, width=\textwidth]{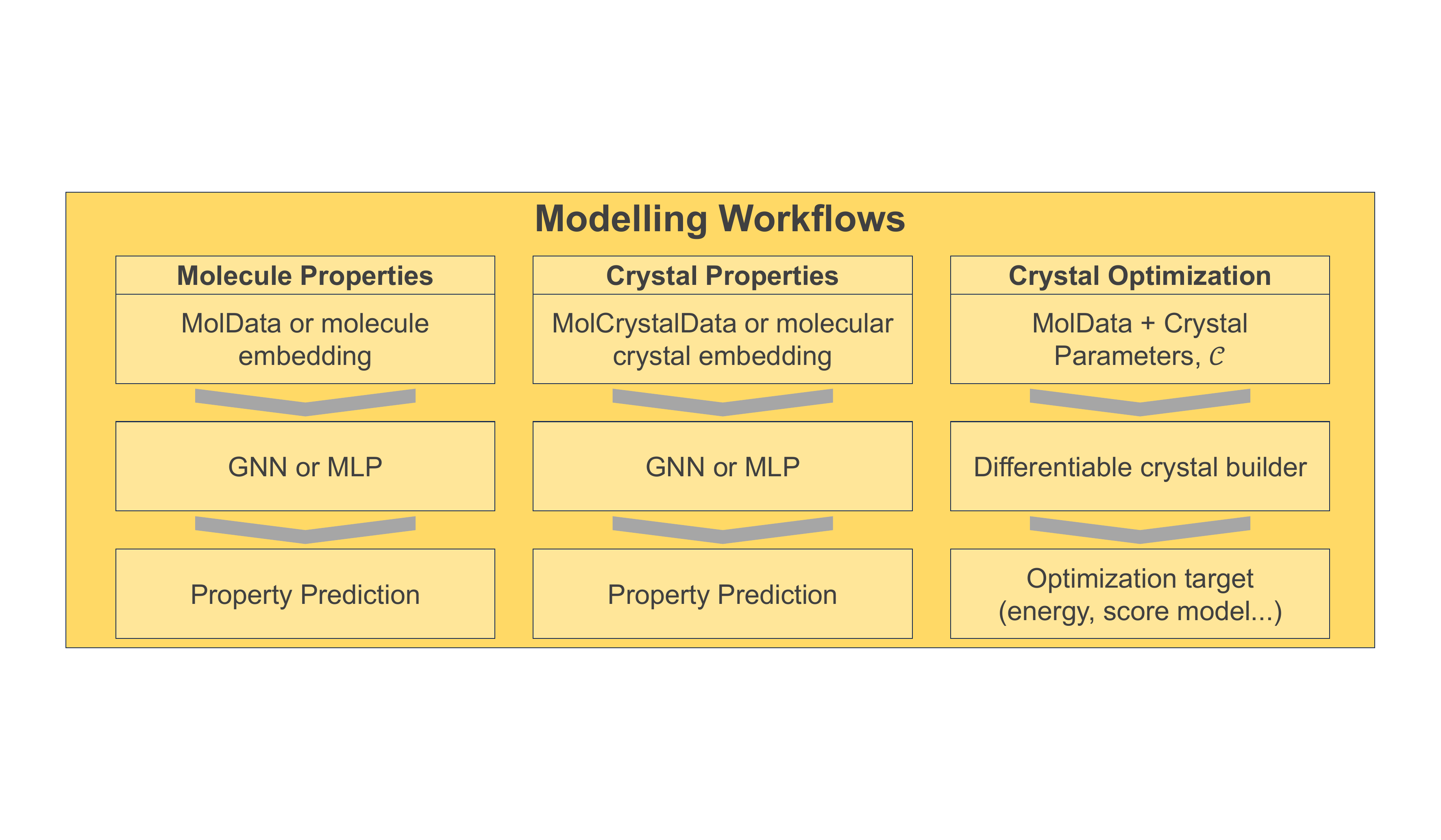}
    \caption{MXtalTools main modelling workflows.
    }
\label{fig:model_workflows}
\end{figure}

We show in Figures~\ref{fig:mol_workflow} and ~\ref{fig:model_workflows} outlines of the workflows for dataset construction and modelling, respectively, including the rules for filtering from cyrstal datasets.

\section{SiLU Potential}

Due to the steep $1/r^{12}$ increase of the standard Lennard-Jones repulsive term at short distances, it can sometimes be an unstable optimization target, especially in situations with large unphysical interatomic overlaps.
We define a simple and robust interatomic potential that roughly maintains the shape and position of the LJ potential well, with linear repulsion for nonzero van der Waals overlaps.
While unphysically soft, this potential never explodes / is always well-behaved, even for poor crystal structures.
Unlike the Buckingham potential, this energy also requires no fine-tuning of constants, and does not diverge near zero.

The SiLU potential, so named from the use of the `Sigmoid Linear Unit' activation function, common in neural network modelling, between two atoms is given as
\begin{equation}
    E_{SiLU}(r)=\frac{7}{25}SiLU\left(-4R\left(r-\sigma R\right)\right),
\end{equation}
for SiLU the logistic sigmoid function, $\frac{x}{1+e^{-x}}$, r the interatomic distance, $\sigma$ the sum of atoms' van der Waals Radii, and $R$ a scaling factor which shifts the repulsive onset to the left (softening) or right (hardening), set to 1 by default.

\begin{figure}
\centering
    \includegraphics[trim=0 0 0 0, clip, width=0.5\textwidth]{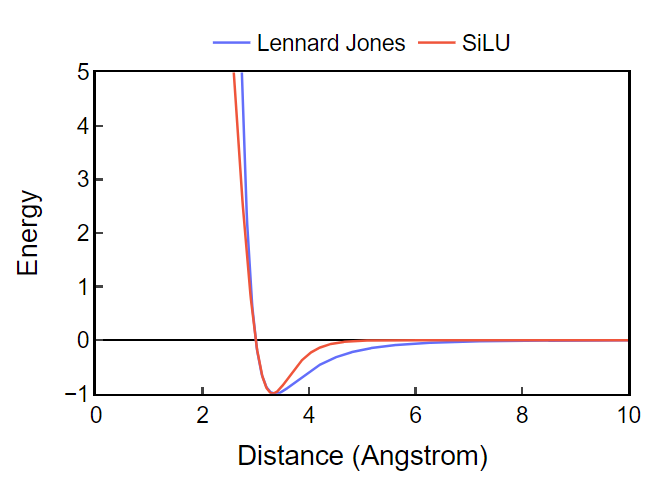}
    \caption{Visual comparison of SiLU and Lennard-Jones potentials near the potential minimum, with $\sigma$ taken as 3 \r{A}, $R=1$.}
\label{fig:silu}
\end{figure}

A visual comparison to the standard Lennard-Jones potential is given in Figure~\ref{fig:silu}.
We see that the location and rough shape of the minima agree, though the SiLU attraction decays faster.
This is actually advantageous in the typical role of this potential, that is, coarse structure optimization, where the shorter range of the potential allows for shorter interatomic cutoffs, 6\r{A} vs. 10 for full LJ.

\section{Crystal Parameterization}

\begin{figure}
\centering
    \includegraphics[trim=0 160 0 80, clip, width=\textwidth]{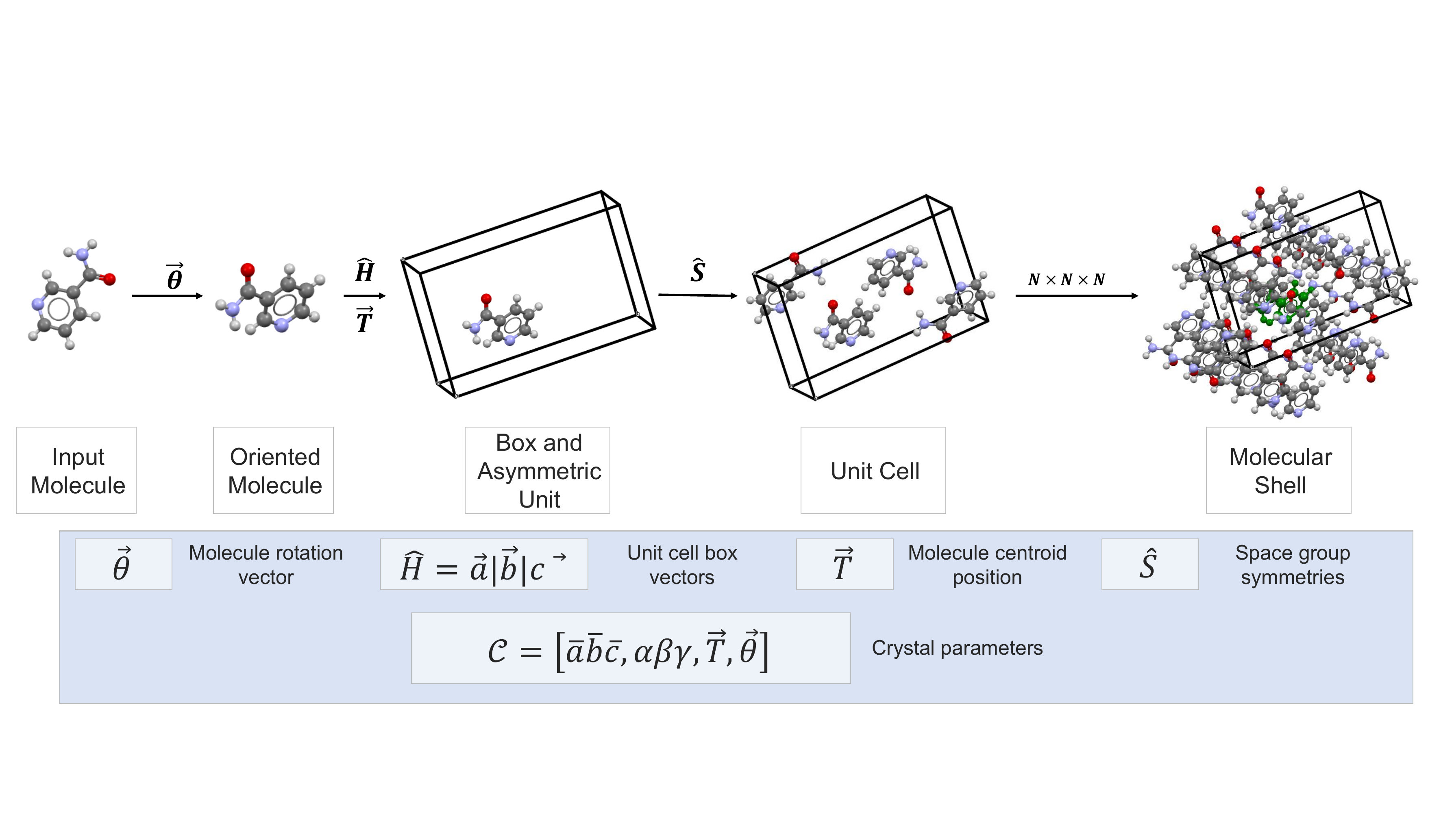}
    \caption{Graphic representation of our crystal building pipeline for a $Z'=1$ molecular crystal, combining the molecule, position and orientation in the asymmetric unit (pose), crystal symmetry operations (space group) and box vectors to generate the unit cell.
    For subsequent analysis, we pattern an $N\times N\times N$ supercell and carve out a cluster surrounding the asymmetric unit.
    }
\label{fig:crystal_pipeline}
\end{figure}
A rigid $Z'=1$ molecular crystal placed on a general Wyckhoff position is completely described by the molecular conformation, space group symmetry, and 12 crystal parameters, $\mathcal{C}$.
These parameters correspond to cell vector lengths ($\bar{a}$,$\bar{b}$,$\bar{c}$), internal angles ($\alpha$,$\beta$,$\gamma$), the position of the molecule centroid in unit cell fractional coordinates $\vec{T}=(u, v, w)$, and the molecule orientation defined against a standardized orientation by a rotation vector $\vec{\theta}=(x, y, z)$.
The box vectors are defined in cartesian coordinates as
\begin{equation}
    \vec{a}=(\bar{a}, 0, 0)
\end{equation}
\begin{equation}
    \vec{b}=(\bar{b}\cos\gamma,\bar{b}\sin\gamma,0)
\end{equation}
\begin{equation}
    \vec{c}=\bigg(\bar{c}\cos\beta, \bar{c}\frac{\cos\alpha-\cos\beta\cos\gamma}{\sin\gamma}, \frac{V}{\bar{a}{\bar{b}\sin\gamma}}\bigg),
\end{equation}
with $V$ the unit cell volume.
The cell parameters and crystal building process is outlined in Figure~\ref{fig:crystal_pipeline}.
Box parameters are generally given explicitly in crystal .cif files, and pose parameters can be extracted using MXtalTools data processing utilities.

\section{COMPACK Parameters}

The following parameters were used for the COMPACK structure comparison in the case study in the main text.

distance tolerance = 0.4

angle tolerance = 40

allow molecular differences = True

packing shell size = 20

\section{Crystal Search}

\begin{figure}  
\centering
    \includegraphics[trim=0 300 0 300, clip, width=1.0\textwidth]{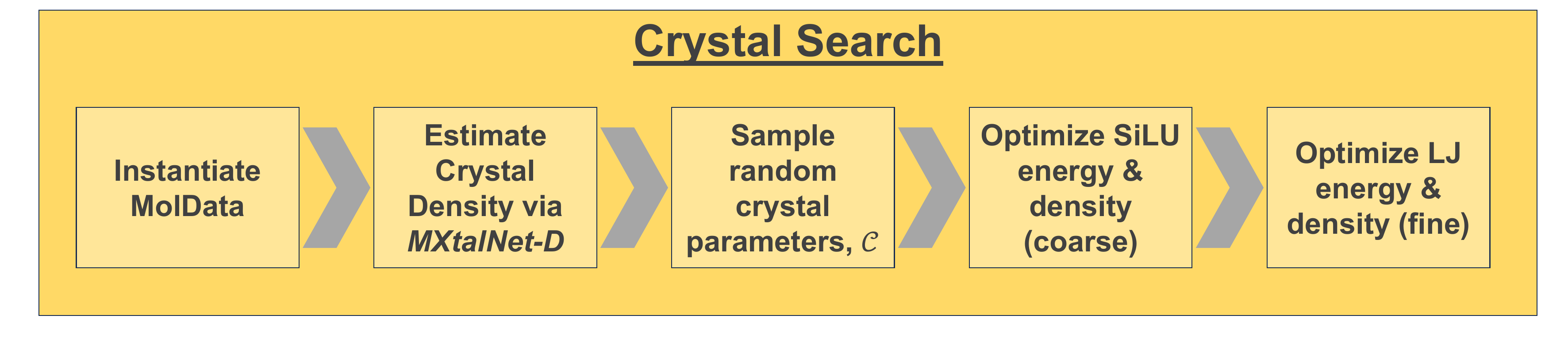}
    \caption{Workflow diagram for our crystal search case study.
    }
\label{fig:csp_workflow}
\end{figure}

The procedure for the crystal search case study follows the workflow outlined in Figure~\ref{fig:csp_workflow}.
The Python code required to run such a search is provided below and the .yaml config file used for the optimization is also provided.

\begin{lstlisting}[language=python]

import os
import subprocess
import sys

import numpy as np
import torch
from tqdm import tqdm

from examples.crystal_search_reporting import batch_compack, density_funnel, compack_fig
from mxtaltools.analysis.crystal_rdf import compute_rdf_distance

# add MXtalTools to path by relative reference
sys.path.insert(0, os.path.abspath("../"))

import torch.nn.functional as F
from mxtaltools.common.training_utils import load_crystal_score_model
from mxtaltools.dataset_utils.utils import collate_data_list
from mxtaltools.models.utils import softmax_and_score

torch.set_grad_enabled(False)

device = 'cuda'
dafmuv_path = "datasets/DAFMUV.pt"
mini_dataset_path = '../mini_datasets/mini_CSD_dataset.pt'
score_checkpoint = r"../checkpoints/crystal_score.pt"
density_checkpoint = r"../checkpoints/cp_regressor.pt"
opt_path = r"opt_outputs/DAFMUV.pt"

"""
Load crystal score model
"""
score_model = load_crystal_score_model(score_checkpoint, device).to(device)
score_model.eval()

"""
Load and analyze reference crystal 
"""
dafmuv_data = torch.load(dafmuv_path, weights_only=False)
ref_crystal_batch = collate_data_list(dafmuv_data).to(device)
ref_computes, ref_cluster_batch = ref_crystal_batch.analyze(
    computes=['lj'], return_cluster=True, cutoff=10, supercell_size=10)
model_output_ref = score_model(ref_cluster_batch.to(device), force_edges_rebuild=True).cpu()
ref_score = softmax_and_score(model_output_ref[:, :2]).cpu()
ref_pred_rdfemd = F.softplus(model_output_ref[:, 2]).cpu()
rdf, bin_edges, _ = ref_cluster_batch.compute_rdf()
ref_rdf = rdf.cpu()

ref_lj_energy = ref_computes['lj'].cpu()
ref_cp = ref_cluster_batch.packing_coeff.cpu()

"""
Run optimization via standalone script & config
"""
subprocess.run(["python", "run_search.py", "--input", "dafmuv_example.yaml"], check=True)

"""
Analyze optimized samples
"""
opt_sample_list = torch.load(opt_path, weights_only=False)
batch_size = 25
num_batches = len(opt_sample_list) // batch_size + int((len(opt_sample_list) % batch_size) > 0)
opt_score, opt_pred_rdfemd, opt_rdfs, opt_lj_energy, opt_cp = [], [], [], [], []
for batch_idx in tqdm(range(num_batches)):
    opt_crystal_batch = collate_data_list(opt_sample_list[batch_size * batch_idx:batch_size * (1 + batch_idx)]).to(
        device)
    computes, opt_cluster_batch = opt_crystal_batch.analyze(
        computes=['lj'], return_cluster=True, cutoff=10, supercell_size=10
    )

    model_output = score_model(opt_cluster_batch.to(device), force_edges_rebuild=True).cpu()
    opt_score.append(softmax_and_score(model_output[:, :2]).cpu())
    opt_pred_rdfemd.append(F.softplus(model_output[:, 2]).cpu())
    rdf, bin_edges, _ = opt_cluster_batch.compute_rdf()
    opt_rdfs.append(rdf.cpu())
    opt_lj_energy.append(computes['lj'].cpu())
    opt_cp.append(opt_crystal_batch.packing_coeff.cpu())

opt_score = torch.cat(opt_score)
opt_pred_rdfemd = torch.cat(opt_pred_rdfemd)
opt_rdfs = torch.cat(opt_rdfs)
opt_lj_energy = torch.cat(opt_lj_energy)
opt_cp = torch.cat(opt_cp)

"""
Compute true RDF distances
"""
rdf_dists = torch.zeros(len(opt_rdfs), device=opt_rdfs.device, dtype=torch.float32)
for i in range(len(opt_rdfs)):
    rdf_dists[i] = compute_rdf_distance(ref_rdf[0], opt_rdfs[i], bin_edges.to(opt_rdfs.device)) / \
                   ref_cluster_batch.num_atoms[0]
rdf_dists = rdf_dists.cpu()

"""
COMPACK analysis
"""
best_sample_inds = torch.argwhere((opt_cp > 0.6) * (opt_cp < 0.8)).squeeze()
matches, rmsds = batch_compack(best_sample_inds, opt_sample_list, ref_crystal_batch)

all_matched = np.argwhere(matches == 20).flatten()
matched_rmsds = rmsds[all_matched]

"""
Figures
"""
good_inds = torch.argwhere(opt_pred_rdfemd < 0.015).flatten()
density_funnel(opt_pred_rdfemd[good_inds],
               opt_cp[good_inds],
               rdf_dists[good_inds],
               ref_pred_rdfemd,
               ref_cp,
               yaxis_title='Predicted Distance',
               write_fig=True)
good_inds = torch.argwhere(opt_lj_energy < -350).flatten()
density_funnel(opt_lj_energy[good_inds],
               opt_cp[good_inds],
               rdf_dists[good_inds],
               ref_lj_energy,
               ref_cp,
               yaxis_title='LJ Energy (Arb Units)',
               write_fig=True)

compack_fig(matches, rmsds, write_fig=True)

\end{lstlisting}

The below file configures the optimization script \verb|run_search.py|, with the target packing coefficient predicted via MXtalNet-D using the above-provided workflow.

\begin{lstlisting}[language=yaml]
device: cuda
mol_path: /crystal_datasets\DAFMUV.pt
out_dir: /opt_outputs
score_model_checkpoint: checkpoints/crystal_score.pt
run_name: DAFMUV

mol_seed: 0
opt_seed: 0
sampling_mode: all
mols_to_sample: 1
num_samples: 1000

sgs_to_search: [33]
zp_to_search: [1]

batch_size: 1000
grow_batch_size: false

init_sample_method: random
init_sample_reduced: falsenon-niggli init sampling
init_target_cp: 0.699 

opt:
  - optim_target: 'silu'
    enforce_niggli: false
    compression_factor: 1.0
    target_packing_coeff: 0.699
    init_lr: 0.001
    convergence_eps: 0.001
    optimizer_func: 'rprop'fastest and most reliable
    anneal_lr: false
    grad_norm_clip: 0.1
    show_tqdm: true
    max_num_steps: 500

  - optim_target: 'lj'
    enforce_niggli: false
    compression_factor: 0.0
    target_packing_coeff: 0.699
    init_lr: 1.0
    convergence_eps: 0.0001
    optimizer_func: 'sgd'
    anneal_lr: true
    grad_norm_clip: 0.01
    show_tqdm: true
    max_num_steps: 50
\end{lstlisting}

\end{document}